# A Memetic Algorithm with Reinforcement Learning for Sociotechnical Production Scheduling

**FELIX GRUMBACH[1], NOUR ELDIN ALAA BADR[1], PASCAL REUSCH.[1] and SEBASTIAN TROJAHN[2]**
[1]Center for Applied Data Science (CfADS), Bielefeld University of Applied Sciences, Gütersloh, Germany
[2]Deptartment of economics, Anhalt University of Applied Sciences, Bernburg, Germany

Corresponding author: Felix Grumbach (e-mail: f.grumbach@ymail.com).

Open Access funding enabled and organized by Projekt DEAL. This work was supported as part of the joint research project *Human-centered Smart Service Lab / Predictive Scheduling* (with project number [EFRE-030018] of the European Regional Development Fund) which is funded by the federated state North Rhine-Westphalia, Germany.

**ABSTRACT** The following interdisciplinary article presents a memetic algorithm with applying deep reinforcement learning (DRL) for solving practically oriented dual resource constrained flexible job shop scheduling problems (DRC-FJSSP). From research projects in industry, we recognize the need to consider flexible machines, flexible human workers, worker capabilities, setup and processing operations, material arrival times, complex job paths with parallel tasks for bill of material (BOM) manufacturing, sequence-dependent setup times and (partially) automated tasks in human-machine-collaboration. In recent years, there has been extensive research on metaheuristics and DRL techniques but focused on simple scheduling environments. However, there are few approaches combining metaheuristics and DRL to generate schedules more reliably and efficiently. In this paper, we first formulate a DRC-FJSSP to map complex industry requirements beyond traditional job shop models. Then we propose a scheduling framework integrating a discrete event simulation (DES) for schedule evaluation, considering parallel computing and multicriteria optimization. Here, a memetic algorithm is enriched with DRL to improve sequencing and assignment decisions. Through numerical experiments with real-world production data, we confirm that the framework generates feasible schedules efficiently and reliably for a balanced optimization of makespan (MS) and total tardiness (TT). Utilizing DRL instead of random metaheuristic operations leads to better results in fewer algorithm iterations and outperforms traditional approaches in such complex environments.

## I. INTRODUCTION
### A. TRADITIONAL JOB SCHEDULING MODELS
Job scheduling is a traditional optimization problem that involves arranging activities in a specific order (sequencing), assigning them to resources with limited capacity, and optimizing one or more objectives [1]. This problem and its various subproblems, methods for solving them efficiently as well as applications, are still of significant scientific interest in several fields. One important subproblem is the job shop scheduling problem (JSSP). Here, a set of jobs is scheduled on a set of machines. Each job consists of a sequence of operations and each operation must be processed on a specific machine for a specific processing time. The machine order can be different for each job, a machine can only process one operation at a time and operations can only start when the predecessor operation in finished. An operation must be processed exactly once and cannot be interrupted. Common optimization objectives for JSSPs are the minimization of MS, TT, flow time or any kind of manufacturing costs such as energy prices. Depending on the objective, the JSSP is an NP-hard sequencing problem [2]. The Flow Shop Scheduling Problem (FSSP) is also a classic optimization problem. It is a specialization of the JSSP, with the difference that all jobs visit the machines in the same order [1]. On the other hand, the flexible JSSP (FJSSP) is a generalization of the JSSP: An operation can be processed on a set of alternative machines, but must be assigned to one machine, which can also affect the processing time [3]. Thus, the FJSSP is not only a job sequencing problem but also a resource assignment (synonym: allocation) problem. If two resource types can be scheduled (e.g. machines and human workers) in a flexible manner, then it is called dual resource constrained FJSSP (DRC-FJSSP) [4].

### B. MOTIVATION AND SCHEDULING REQUIREMENTS IN PRACTICE

The study was motivated by a lack of efficient and holistic scheduling models for flexible and human-centered production and assembly processes, which are common in manufacturing companies. Considering requirements from our industrial partners, this is particularly relevant for sociotechnical, discontinuous manufacturing environments with high-mix low-volume products. The following aspects must be taken into account: Setup and processing operations can be performed by human workers with different capabilities, where the worker training level affects the operation feasibility and duration. In common practice, setup and processing operations are sometimes carried out by different workers. For example, a trained specialist sets up a machine and an assistant then operates it. Furthermore, workstations can have different characteristics: Some stations require setup operations (e.g. machinery and equipment) and others do not (e.g. assembly workplaces). The setup duration depends not only on the worker's capability but also on the current setup status (sequence-dependent setup time). For example, a machine must be extensively retooled if two very different components are sequentially processed. On the other hand, if two very similar components are manufactured one after the other on the machine, the setup only involves a small change of the machine setting. In some cases, machines can perform work (partially) autonomously. These are, for example, computer-aided machine tools or industrial robots. In series production, it can often be observed that a worker interacts with several machines at the same time and only loads or re-adjust the machines. Some workstations have multiple slots that can be worked on at the same time (e.g. assembly workbenches). There are also company-external stations for outsourced work processes (extended workbenches). In this context, transport logistics tasks are also required. Moreover, sequential jobs in conventional models such as JSSP or FJSSP are insufficient for practical requirements on more complex structures. This circumstance becomes transparent in the course of the BOM explosion: Manufacturing multi-part products require highly structured jobs with dependencies and the possibility of parallelization. Given this fact, jobs should be mapped as directed acyclic graphs (DAG), which is comparable with the resource-constrained project scheduling problem (RCPSP) nature [5].

### C. METAHEURISTICS

Due to the ever more complex scheduling models, various heuristic approaches have been presented in the past years. This especially includes metaheuristics with population-based methods such as genetic algorithms (GA), methods with a single starting solution (trajectory methods) such as simulated annealing (SA) or hybrid methods such as memetic algorithms. A major benefit of using population-based metaheuristics like GA is the ability to obtain reliable results [6]. Another advantage of metaheuristics is that they can be implemented loosely coupled [7]. As a consequence, changes to the environment or to the evaluation function do not necessarily require changes to the optimization algorithm operations. Moreover, this can be combined with the paradigm of simheuristics, where a simulation model is integrated into an optimization method to generate or analyze solutions [8]. Simulation models are well suited to depicting complex scenarios and are therefore open to the expansion of further features and constraints [9]. Compared to mathematical models, they can be analyzed and developed more intuitively.

### D. DRL

In recent years, DRL has increasingly attracted the interest of researchers. DRL is a machine learning technique for solving optimization problems with deep neural networks (DNN). Based on conditioning, a DRL agent learns a policy and how to act in a virtual environment in such a way to maximize rewards. The method can be summarized as a Markov Decision process: A DRL agent receives a time discrete environment state. Afterward, the agent has a set of actions available, that can be performed. Executing an action manipulates the environment state and leads to a direct or delayed distribution of a reward or punishment defined by a reward function. The DNN is used as a function approximator to predict promising actions for respective environment states [10]–[12]. DRL achieves high-quality results with very short computing times, provided the model is well trained [13]. However, under- or overfitted models lead to poor performance [14], [15]. Especially in realistic environments with demand and capacity fluctuations, novel situations, large product ranges and changing production goals, this can lead to suboptimal results. This may be one reason that most authors of pure DRL approaches are limited to simple environments such as JSSP or FJSSP.

### E. CONTRIBUTION OF THIS WORK

From the current literature, we note a lack of holistic models combined with efficient algorithms for sociotechnical production scheduling evaluated with real-world data (see Section II for the complete definition of the research gap). To fulfill this gap, the paper provides the following contributions:

1) A mathematical model that maps the domain constraints and requirements mentioned (see Section III).
2) An efficient scheduling framework, where a DRL agent is injected into a GA to improve sequencing and assignment decisions (see Section IV). In this way, we combine the mentioned advantages of metaheuristics and DRL to obtain better and more reliable results.

With the scheduling framework, the following practically relevant aspects are considered:

- A generic and customizable DES is used by the optimization algorithm to generate and evaluate schedules.
- The framework supports parallel computing and can therefore be used scalably in a cloud.
- Objectives can be flexibly defined based on simulation metrics. In this work, MS and TT are exemplary balanced.

In Section V, numerical experiments including benchmarks with self-generated and real-world data are carried out and discussed. It was confirmed that DRL improves the performance of a memetic algorithm by obtaining better results in fewer iterations, which is promising for reactive real-time use.

## II. RELATED WORK

The following section gives an overview of recent studies in the area of job scheduling models for flexible manufacturing processes and highlights research gaps. In particular, representative metaheuristics, simheuristics and DRL approaches for FJSSP and DRC-FJSSP have been considered.

### A. SOLVING FJSSP

Winklehner and Hauder [16] reviewed the previous scientific literature and concluded that no previous work considered release dates, deadlines and sequence-dependent setup times in FJSSP, especially in real-world problems. Thenarasu *et al.* [17] proposed a solution for FJSSP using composite dispatching rules and multi-criteria decision making based priority rules integrated with DES. They tested the solution on real-world data. Zhang *et al.* [18] proposed a GA to solve an extended FJSSP with multiple sequence-dependencies for setup and transportation times. Moreover, three objectives have been considered: MS minimization, total setup time minimization, and transportation time minimization. Zhu and Zhou [19] examined solving a FJSSP with job precedence for MS minimization. In this way, complex job paths, as they arise in the production of BOM components, could be modeled. Depending on the structure of the input data, it is also possible to schedule parallel operations. A multi-micro-swarm leadership hierarchy-based optimization algorithm has been carried out to solve this problem. Lunardi and Voos [20] considered parallel operations within DAG job paths. The authors presented a GA and a firefly algorithm for MS minimization.

Defersha and Rooyani [21] evaluated the effects of using parallel computation on large FJSSP problem instances. They imposed sequence-dependent setup times and used a GA to minimize the MS. It was shown that parallel computing significantly improves the computing time for population-based methods. Furthermore, Huang and Yang [22] offered important insights into using hybrid algorithms such as GA and SA for improving the results. Here, a multicriteria objective was used to balance MS, workload of critical machines, and total workloads of machines. Sequence-dependent transportation times were taken into account.

### B. SOLVING DRC-FJSSP

In recent years, DRC-FJSSP has been considered more frequently to schedule additional resources. Wu *et al.* [23] considered human workers and their learning ability over time. They combined a GA with variable neighborhood search for MS minimization. Similar to this approach, Du *et al.* [24] presented a dual resource-constrained flexible flow shop scheduling problem with human workers and the consideration of human fatigue. They proposed a GA-based hybrid metaheuristic to optimize the MS. Defersha *et al.* [25] considered detached setup operations, thus they proposed DRC-FJSSP with machines and setup operators constraints. Furthermore, they took into account sequence-dependent setup times and workload balancing. They developed an SA algorithm to solve the problem. Liu *et al.* [26] investigated multi-worker assignments and parallel team scheduling for a product assembly line scheduling problem. A simultaneous optimization of MS and imbalance degree of team workload is taken into account, which has been applied to real-world data. Berti *et al.* [27] proposed a model, where the age of workers has an affect on their experience level, physical capacity and rest allowance. Furthermore, they demonstrated the importance of balancing intensive work and resting times, taking into account the age of workers. Dispatching rules were used for the scheduling approach: The shortest processing time (SPT) rule was applied for non-scheduled breaks to get a lower MS, while the longest processing time (LPT) rule was applied for scheduled breaks. Recently, there has been an increased emphasis on solving DRC-FJSSP with more complexity, such as multiple objectives and sequence-dependencies. Zhang and Jie [28] attempted to solve a DRC-FJSSP with multiple objectives by balancing MS and production costs (minimizing stocking of raw materials, machines, worker costs and tardiness penalties). Andrade-Pineda *et al.* [29] dealt with machine-related skills and suitabilities of workers. Moreover, dynamic due dates and multiple objectives such as MS minimization and weighted tardiness minimization are taken into account. Kress *et al.* [30] presented a DRC-FJSSP with flexible human workers in which sequence-dependent setup times are taken into account. Here, the machine setup state has an effect on the duration of the next setup operation.

### C. APPLYING DRL

In the recent past, some scholars examined how DRL can be applied to scheduling problems such as the FJSSP: Lang *et al.* [31] evaluated the effects of integrated DRL agents for assignment and sequencing decisions with a DES. The DES was used to execute the agent's decisions and to ob-serve objectives such as MS or TT. Numerical experiments showed that DRL delivers high-quality results in near real time. Khuntiyaporn et al. [32] focused on solving FJSSP using DRL. They simulated each machine into two sets of machines with different processing times and costs to solve a single objective such as MS minimization. Popper et al. [33] applied DRL in FJSSP for multicriteria optimization. Multiple DRL agents were used to minimize MS, delayed products, lateness and energy costs. In order to examine a FJSSP with dynamic disturbances such as new job insertion and machine disturbance, they tested their approach with experiments in a simulation. Luo et al. [34] proposed a hierarchical multiagent DRL algorithm named hierarchical multi-agent proximal policy optimization to select the best fitting dispatching rules for sequencing and assignment decisions dynamically. The algorithm aims to optimize total weighted tardiness, average machine utilization and variance of machine workload as multiple objectives. Zhou et al. [35] proposed a DRL smart scheduler that handles unexpected events to solve JSSP, besides they considered multicriteria optimization by optimising composite reward functions. Du et al. [36] considered solving a FJSSP with setup operations, machine processing and transportation times to minimize both MS and total electricity costs. The problem was solved by applying an estimation of distribution algorithm for exploration while DQN was applied for better exploration. Zhu et al. [37] introduced a multi-agent DRL algorithm that solves DRC-FJSSP in real-time. Three agents are trained to perform three decision-making tasks: process planning, job sequencing, and machine selection. Moreover, different dispatching rules are set for each decision-making task to fulfill the requirements of the action space in the DRL algorithm. Furthermore, they considered DAG for defining the precedence relationships of operations.

### D. COMBINING DRL AND METAHEURISTICS

Recent studies have also combined metaheuristic algorithms with DRL. Seyyedabbasi et al. [38] compared three metaheuristics models combined with DRL. They concluded that combining reinforcement learning with metaheuristics leads to a more stabilized performance in exploration and exploitation when compared to sole metaheuristic models. Thus, the model aims to discover additional areas and avoids being trapped in local optima, which enables the model to find better solutions. Kosanoglu et al. [39] combined a metaheuristic approach with DRL. They combined SA with a double deep Q network. In their approach, the initial solution for the SA model is not generated randomly but obtained from the best solution provided by the DRL algorithm. The best solution obtained from the SA model is provided as an initial solution for the DRL algorithm. They deduced that this hybrid algorithm obtained the optimal solution and is capable of outperforming other metaheuristic algorithms without combining with DRL.

### E. RESEARCH GAP

Based on the literature reviewed, a research gap can be identified: There is a lack of holistic mathematical models and approaches to generate schedules, that fully provide the introduced requirements [36], [40], [41]. Several recent studies have already examined and modeled partial aspects, but not combined them. Most authors used population-based or hybrid methods, which may indicate the good suitability of this class of algorithms. However, DRL-based approaches are still primarily restricted to simple environments. Further work is required to investigate the applicability of DRL, especially in the context of DRC-FJSSP and how to reliably generate schedules in this way. It should also be emphasized, that DES models and the evaluation with real-world data have been rarely utilized. Especially, due to the highly complex reality, modern process simulation tools allow more open and simple modeling compared to pure mathematical models. We consider it very relevant to separate a simulation from the optimization procedure as much as possible to allow for customizing and extensions. Moreover, we could only identify Defersha and Rooyani [21], who optimized the computing time through parallel computing. Computational efficiency is an important requirement for a time-sensitive reactive deployment with fast re-scheduling. Table 1 classifies the studies investigated and shows a clear gap in this context.

TABLE 1: Related work matrix of the reviewed literature. For a better overview, the studies are clustered according to their content. Includes the contribution of the current paper. (MRS: Multi resource scheduling, MCO: Multi criteria optimization, SD: Sequence-dependencies, PO: Parallel operations (=complex BOM job paths), Sim: Simulation/Simheuristic, RWD: Real-world data)

| Study | MRS | MCO | SD | PO | Sim | DRL | RWD |
|---|---|---|---|---|---|---|---|
| [21] | | | • | | | | |
| [18] | | | • | | | | |
| [22] | | • | • | | | | |
| [25] | | • | • | | | | |
| [16] | | | • | | | | • |
| [17] | | • | | | • | | • |
| [26] | • | • | | | | | • |
| [30] | • | | | | | | • |
| [23] | • | | | | | | |
| [27] | • | | | | | | |
| [24] | • | | | | | | |
| [28] | • | • | | | | | |
| [29] | • | • | | • | | | |
| [37] | • | | | | | • | |
| [31] | | | | | • | • | |
| [32] | | • | | | • | • | |
| [33] | | • | • | | | • | |
| [36] | | • | • | | | • | |
| [34] | | • | | | | • | |
| [35] | | • | | | | • | |
| [20] | | | | • | | | |
| [19] | | | | • | | | |
| Contribution of this paper | • | • | • | • | • | • | • |

## III. PROBLEM FORMULATION

Considering the practical requirements introduced, we formulate the following composed mixed integer linear program (MILP) that goes beyond traditional scheduling models. It corresponds to a DRC-FJSSP with flexible stations, infinite station buffers, flexible human workers, different worker capabilities with an effect on setup and processing times, DAG job paths for BOM manufacturing, (partially) automated tasks, task release times for raw material arrivals, sequence-dependent setup times and early (left-shifted) setup operations. With early setup operations machines can be set up before materials are available which are required for the processing operation. The MILP enables a simultaneous optimization of machine assignments, worker assignments and task sequences. Furthermore, setup operations with sequence-dependent setup times are also taken into account. This enables the optimization of short-term lot sizes by aggregating processing tasks with the same or similar setup operations on a station. To the best of our knowledge, this combination of constraints has not yet been modeled in the scientific literature. However, these are relevant to map the circumstances in medium-sized sociotechnical production processes as in the case with our industrial partners.

TABLE 2: Composed MILP notation

| Notation | Description |
|---|---|
| $I$ | Set of all job indices |
| $T_i$ | Set of all tasks indices of a job $i$ |
| $X$ | Set of all setup and processing operations (see Eq. 1) |
| $K$ | Set of all station indices |
| $W$ | Set of all worker indices |
| $L$ | Matrix over all operation combinations ($L = X \times X$) |
| $L_k$ | Matrix over all operation combinations, that can be processes on station k |
| $r_{i,j}$ | Release time (raw material arrival time) of a task $(i,j)$. This just refers to the processing operation. Setup operations can start earlier. |
| $f_{i,i'}$ | 1, if job $i$ succeeds job $i'$. Else 0 |
| $c_i$ | Job due date. Only possible for jobs without successor jobs. |
| $l_i$ | Last task of a job (index) |
| $M_{i,j}$ | Set of alternative stations for a task |
| $N_{i,j,s,k}$ | Set of alternative workers for a processing or setup operation $(i,j)$ on a station $k$ (defined by capability matrix) |
| $O_{i,j,s}$ | Set of alternative workers, that can execute a processing or setup operation $(i,j,s)$ |
| $P_k$ | Set of task tuples $(i,j)$ that can be processed on station $k$ |
| $Q_{i,j,s,w}$ | Set of stations, where a worker $w$ can execute a specific operation $(i,j,s)$ |
| $d_{i,j,s,k,w}$ | Setup or processing operation $(i,j,s)$ duration by worker $w$ on station $k$ |
| $s_{i,j,i',j',k}$ | Sequence-dependency factor $\in [-1, 0.5]$ to calculate the setup time, if task $(i,j)$ preceeds $(i',j')$ directly on station $k$. |
| $u_{i,j,k}$ | Automation degree $\in [0,1]$ of a processing operation $(i,j)$ |
| $w_1, w_2$ | Objective weights $\in [0,1]$ to balance MS ($w_1$) and TT ($w_2$) optimization |
| $\alpha_{i,j,s}$ | Start time of an operation (variable) |
| $\beta_{i,j,s}$ | Completion time an operation (variable) |
| $\sigma_i$ | Tardiness of job $i$ (variable) |
| $\gamma, \delta, \lambda, \psi$ | Binary assignment and sequencing variables (see text) |

To fully depict the complex DAG paths, the model has a three-level job hierarchy: A job (denoted by $i$, 1st level) has several tasks ($j$, 2nd level) and a task has a setup and processing operation ($s$, 3rd level). This is defined over the set $X$ in Eq. 1, which contains all operation triples.

$$X = \left[(i,j,s) \mid i \in I, j \in T_i, s = \begin{cases} 1, \text{if it is a setup operation} \\ 0, \text{else} \end{cases}\right] \quad (1)$$

Table 2 provides the model parameters and the time variables $\alpha \in \mathbb{R}_{\geq 0}, \beta \in \mathbb{R}_{\geq 0}, \sigma \in \mathbb{R}_{\geq 0}$. In addition, four binary variables $\gamma, \delta, \lambda, \psi$ are used to assign and sequence the tasks. We know that the number of variables can be reduced, especially when the problem is modeled as a mixed integer non-linear program (MINLP). However, for better comprehension and readability of the constraints, we rely on the auxiliary variables in this section. $\gamma$ is an assignment variable that determines if a task is allocated to a specific station from a set of alternative stations. The choice of station can have an impact on the processing or setup time. For example, a modern machine could be set up faster and also process faster.

$$\gamma_{i,j,k} = \begin{cases} 1, \text{if the task } (i,j) \text{ is processed on station } k \\ 0, \text{else} \end{cases} \quad (2)$$

Depending on which worker from a set of alternative workers carries out a setup or processing operation, this can also have an effect on processing or setup times. For example, an experienced operator could be faster than an apprentice. $\delta$ is a binary assignment variable to determine the worker to perform an operation on a station. With $d_{i,j,s,k,w}$, the required time to perform this operation can thus be derived.

$$\delta_{i,j,s,k,w} = \begin{cases} 1, \text{if the operation } (i,j,s) \text{ is processed on station } k \\ \quad \text{by worker } w \\ 0, \text{else} \end{cases} \quad (3)$$

$\lambda$ is a sequencing variable to define the structure of the task DAG (see Sect. IV-B). In contrast to most traditional models, an operation can have any number of predecessor and successor operations, which comes closer to realistic circumstances.

$$\lambda_{i,j,s,i',j',s'} = \begin{cases} 1, \text{if the operation } (i,j,s) \\ \quad \text{preceeds operation } (i',j',s') \\ 0, \text{else} \end{cases} \quad (4)$$

$\psi$ is a sequencing variable to determine the sequence-dependent setup time of an operation, taking into account the setup state of a machine and the worker assigned.

$$\psi_{i,j,s,i',j',s',k,w} = \begin{cases} 1, \text{ if operation } (i,j,s) \\ \quad \text{ preceeds operation } (i',j',s') \\ \quad \text{ directly on station } k \text{ and if } (i',j',s') \\ \quad \text{ is done by worker } w \\ 0, \text{ else} \end{cases} \quad (5)$$

Two objective functions are defined to minimize MS and TT synchronously:

$$\min C_{max} = \max_{(i,j,s) \in X} \{ \beta_{i,j,s} \} \quad (6)$$

$$\min T = \sum_{i \in I \mid c_i \text{ is set}} \sigma_i \quad (7)$$

The following constraints ensure that the tardiness is calculated correctly and always positive:

s.t.
$$\sigma_i \geq \beta_{i,l_i,0} - c_i \quad \forall i \in I \mid c_i \text{ is set} \quad (8)$$
$$\sigma_i \geq 0 \quad \forall i \in I \mid c_i \text{ is set} \quad (9)$$

It must be ensured that processing operations can only start, when the assigned raw material is available:

$$\alpha_{i,j,0} \geq r_{i,j} \quad \forall i \in I, \forall j \in T_i \quad (10)$$

Moreover, it must be ensured that all processing and setup operations have the correct duration, considering the human worker assigned (Eq. 11) as well as the station and sequence-dependency (Eq. 12):

$$\beta_{i,j,0} - \alpha_{i,j,0} = \sum_{k \in M_{i,j}} \sum_{w \in N_{i,j,0,k}} \delta_{i,j,0,k,w} * d_{i,j,0,k,w} \quad (11)$$
$$\forall i \in I, \forall j \in T_i$$
$$\beta_{i,j,1} - \alpha_{i,j,1} = \sum_{k \in M_{i,j}} \sum_{w \in N_{i,j,1,k}} \delta_{i,j,1,k,w} * d_{i,j,1,k,w} \quad (12)$$
$$+ \sum_{k \in M_{i,j}} \sum_{w \in N_{i,j,1,k}} \sum_{i' \in I} \sum_{j' \in T_{i'}} \psi_{i',j',0,i,j,1,k,w}$$
$$* d_{i,j,1,k,w} * s_{i',j',i,j,k} \quad \forall i \in I, \forall j \in T_i$$

The following constraints guarantee, that all operations are assigned to valid stations (Eq. 13) and workers (Eq. 14). This includes the station, the setup worker and the processing worker.

$$\sum_{k \in M_{i,j}} \gamma_{i,j,k} = 1 \quad \forall i \in I, \forall j \in T_i \quad (13)$$

$$(\sum_{w \in N_{i,j,1,k}} \delta_{i,j,1,k,w}) - \gamma_{i,j,k} + (\sum_{w \in N_{i,j,0,k}} \delta_{i,j,0,k,w}) \quad (14)$$
$$- \gamma_{i,j,k} = 0 \quad \forall i \in I, \forall j \in T_i, \forall k \in M_{i,j}$$

Furthermore, a human worker can only be utilized to a maximum of 100% at a time. So a worker can handle several partially automated operations in parallel. This is a classic case in practice when a machine operator monitors several machine tools at a time:

$$1 + \lambda_{i,j,s,i',j',s'} + \lambda_{i',j',s',i,j,s} \quad (15)$$
$$- \sum_{k \in Q_{i',j',s',w}} \delta_{i',j',s',k,w} * \begin{cases} u_{i',j',k}, \text{ if } s' = 0 \\ 1, \text{ else} \end{cases}$$
$$\geq \sum_{k \in Q_{i,j,s,w}} \delta_{i,j,s,k,w} * \begin{cases} u_{i,j,k}, \text{ if } s = 0 \\ 1, \text{ else} \end{cases}$$
$$\forall (i,j,s,i',j',s') \in L, \forall w \in O_{i,j,s} \cap O_{i',j',s'}$$

Since stations can have multiple workplaces (slots), it must be ensured that a station can process only one operation per slot at a time:

$$1 + \lambda_{i,j,s,i',j',s} + \lambda_{i',j',s',i,j,s} - \gamma_{i',j',k} \geq \gamma_{i,j,k} \quad (16)$$
$$\forall (i,j,s,i',j',s') \in L, \forall k \in M_{i,j} \cap M_{i',j'}$$

Eq. 17 ensures, that a processing operation cannot start before the corresponding setup operation is completed. Moreover, processing operations of a job cannot overlap in time (Eq. 18). The processing operations of a job cannot start before all predecessor jobs have been completed (Eq. 19).

$$\alpha_{i,j,0} \geq \beta_{i,j,1} \quad \forall i \in I, \forall j \in T_i \quad (17)$$
$$\alpha_{i,j,0} \geq \beta_{i,j-1,0} \quad \forall i \in I, \forall j \in T_i \mid j > 1 \quad (18)$$
$$\alpha_{i,1,0} \geq \beta_{i',l_i,0} \quad \forall i, i' \in I \mid f_{i,i'} = 1 \quad (19)$$

A setup operation must be followed by the corresponding processing operation on the station. However, they do not have to follow one another immediately, so that left-shifted setups are possible even before the job starts or the material is available.

$$\gamma_{i,j,k} - \sum_{w \in N_{i,j,0,k}} \psi_{i,j,1,i,j,0,k,w} = 0 \quad (20)$$
$$\forall i \in I, \forall j \in T_i, \forall k \in M_{i,j}$$

The best found solution $\pi$ is afterward selected from the Pareto front $PF$:

$$\pi = \min_{\pi' \in PF} Z_{\pi'} \quad (21)$$

$Z$ is the linearly scalarized value of a solution taking into account freely definable weights $w_1, w_2$:

$$Z = w_1 * (C_{max})_{normalized} + w_2 * T_{normalized} \quad (22)$$

## IV. PROPOSED SCHEDULING FRAMEWORK

In this section, framework components are described successively.

## A. SCHEDULING FRAMEWORK OVERVIEW

The framework provides a memetic algorithm integrating DRL and a DES to evaluate the schedules under the consideration of complex constraints. Figure 1 displays the main building blocks and their relationship within the scheduling dataflow. The memetic algorithm is based on a GA to control the overall optimization process and to provide a breadth-first search for assignment and sequencing decisions. In addition, a SA procedure is used to further exploit solutions within the GA search strategy (see Section IV-D). In order to evaluate the solutions created by the memetic algorithm operators, the framework integrates a process-based DES (see Section IV-E). Moreover, by integrating a properly trained DRL agent to the DES, sequencing and assignment decisions were further improved while simulating (see Section IV-F). To enhance performance in a scalable manner, the computations of the algorithm can be distributed across a freely definable number of parallel processes within the computational environment.

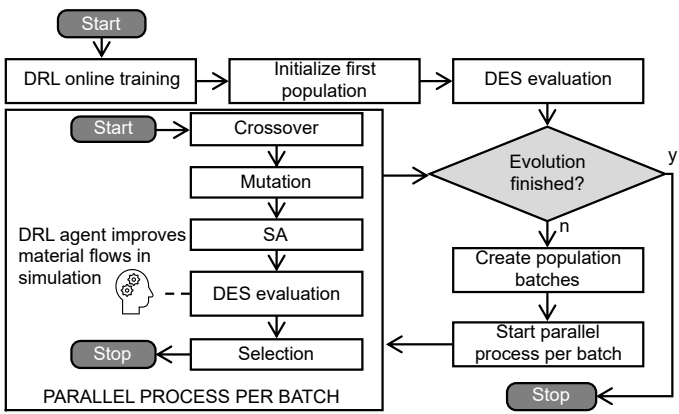

FIGURE 1: Scheduling framework components and dataflow (highlevel representation)

The dataflow proceeds as follows: First, the DNN for the DRL agent is (re-)trained optionally to learn the specifics of the input data (see Section V-B). Then, an initial population of genomes with random assignment and sequencing decisions is created (see Section IV-C). The genomes are handed over to the simulation, improved by DRL and evaluated with regard to their fitness. The algorithm is then executed for a finite number of generations or evaluations (stopping criterion). In each generation, the current base population is partitioned into a number of batches according to the number of parallel computing processes. In this way, it is possible to distribute the computationally intensive simulation workload that take place in each algorithm generation for each genome. Subsequently, the processes are spawned, each iteratively processing the assigned batch step by step. Within each parallel computing process, parent genomes from the batch are crossed over to a given probability. In addition, parent and child genomes can also mutate by a given probability. Afterward, the fittest individual per generation can be further improved by SA for a given probability. At this point it should be mentioned that SA can be exchanged with other trajectory optimization methods. After the simulation-based evaluation, the fittest solutions are selected, from which the next base population emerges. This process repeats until the stopping criterion is reached.

## B. GENOME ENCODING

Due to the architecture of the memetic algorithm, we use a newly developed value encoding technique for the individuals, where a genome consists of two subgenomes to determine the schedule to be evaluated. As shown in Figure 2, a genome is splitted into a resource allocation subgenome and a dispatching subgenome. The resource allocation subgenome consists of resource allocation genes (represented as columns in the figure). Every gene corresponds to a task referring a unique task index and defines the assigned station and workers for the setup and processing operations, which are represented by unique indices $k \in K, w \in W$ (see Table 2). Furthermore, task predecessor/successor relationships are defined by topology groups. A topology group $G \in \mathbb{N}_{>0}$ defines the positional group of a task in a topologically sorted task DAG. This is a required modeling technique to depict complex job paths for BOM manufacturing. The subset of task vertices from all tasks $V$ without incoming edges is located in the first group $G_1$. The subsequent groups $G_{2...n}$ can be determined recursively:

$$G_i = \{v \in V \setminus V^* \mid \forall p \in P_v : p \in V^*\} \quad (23)$$

$V^*$ is the set of all visited vertices. $P_v$ is the set of all direct predecessor vertices of a vertice $v$.

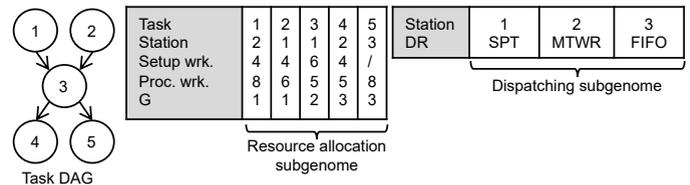

FIGURE 2: Value encoded genome for DAG job paths and to assign setup and processing workers to various stations. (wrk: worker, proc: Processing, DR: Dispatching Rule, SPT: Shortest Processing Time, MTWR: Most Total Work Remaining, FIFO: First In First Out)

As an example from Figure 2, task 3 (topology group 2) is a successor to tasks of topology group 1 (tasks 1 and 2). Following the same concept, it is also a predecessor of all tasks from topology group 3 (tasks 4 and 5). The dispatching subgenome defines selected dispatching rules for stations, where each station is represented by a separate dispatching gene. Enabled by this value encoding technique, flow production systems can also be depicted in this way by fixing the respective station with FIFO. Therefore, this genome coding allows generic applicability to (permutation) flow shops, (flexible) job shops and dual resource flexible job shops according to the formulated model. During DES-based evaluation of genomes via schedule generation (see Section

IV-E), either the preset dispatching subgenome is used, or a DRL agent selects the dispatching rules dynamically for each station (see Section IV-F).

### C. INITIALIZATION

To create a first population as a suitable starting point for the algorithm, a specific initialization technique is implemented with the consideration of the proposed genome encoding. First, a random task DAG is generated for each individual, which ensures a valid topological order of the tasks taking into account the predecessor and successor constraints for BOM production (see Section IV-B). At this stage, the DAG is still independent of any resource allocations. Furthermore, the sequence in which the tasks are processed on a station is also not determined at this stage. Afterward, one gene is created in the resource allocation subgenome, for each node of the DAG. Thereafter, a station and one worker for the processing operation and another for the setup operation are selected from a set of available alternative resources while taking the capability matrix into account (see Table 2). In this case, resources are not randomly selected, rather they are selected in a way that provides an evenly distributed workload. In the resource allocation subgenome, a random dispatching rule is assigned to each station. This determines the sequences of how the tasks are processed on the station. Further rules can be selected in addition to the rules listed in Section IV-F. The detailed procedure can be found in the publicly available source code.

### D. OPERATORS

The following operations are sequentially applied per subgenome couple: First, a Job Order Crossover (JOX) operation (see [42]) is applied to the resource allocation subgenome. We implemented the operator as follows:

1) Randomly select two parent genomes to be crossed.
2) Randomly select 1 to $n-1$ genes from the first parent, where $n$ is the number of all genes.
3) Copy these genes to the same positions in the offspring genome.
4) Select the genes of remaining tasks from the second parent.
5) Copy these genes to the offspring genome in the order in which they appear in the parent.

The dispatching subgenome is inherited from the first parent. Then the mutation is performed, which includes two different options. With a probability of 50%, a resource flip is conducted for a random gene in the resource allocation subgenome. Here, an assigned station, setup worker and processing worker is replaced by a random valid alternative as defined by $M_{i,j}$ and $N_{i,j,s,k}$ (see Table 2). If the resource flip is not conducted, a dispatching rule flip is conducted for a randomly selected gene in the dispatching subgenome. Here, an assigned dispatching rule is replaced by a random valid alternative. In addition to crossover and mutation, a depth-first search is also possible as an additional operator. Within a trajectory method such as SA, a neighborhood search is carried out for the genome. The method receives the fittest genome of a generation and attempts to further optimize it step by step using the mutation operation.

### E. DES MODEL FOR EVALUATION

To decode a genome and calculate its fitness, a DES with several event-based processes is utilized. These are implemented using the *SimPy* framework, which is a leightweight and efficient Python framework to develop DES models with the help of asynchronous generator functions (see [43]). Each station specified in the dispatching subgenome spawns its own station process (see Figure 3). Initially, when no task has yet been assigned, the station is in state *Free*. Then the station sorts its wait queue and pulls an available task (*Dispatching*). All tasks assigned according to the resource allocation subgenome are part of the station queue. For sorting, a dispatching rule is applied, as it is either defined in the dispatching subgenome or as the agent predicts it situationally (see Sec. IV-F). The next step is to decide whether the station needs to be set up or not. In both cases, a separate work process is started (*Setup* or *Processing*). Setup operations can already start independently of the job sequence and the readiness of the associated processing operation. A sequence-dependent set-up time is also taken into account. The worker defined in the assignment gene is awaited in the work process. After the processing operation is completed, the station changes back to the initial state of *Free*. When no more outstanding tasks are available, the station process ends. Different logistic metrics are measured during a simulation run. This includes, for example, MS, TT, flow time or wait times. A free selectable set of normalized metrics can be used in a weighted objective function to determine the fitness.

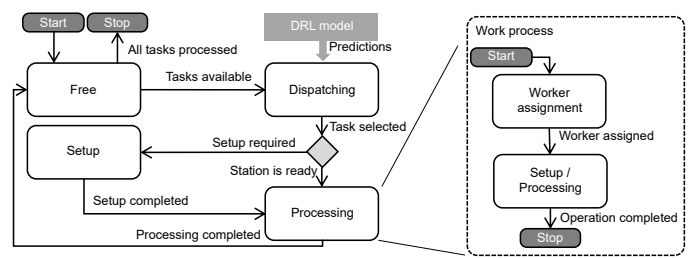

FIGURE 3: State machine for a station simulation process (simplified representation)

### F. DRL IMPLEMENTATION

The DRL agent is injected into the simulation model and improves the genome during the decoding process. First, the DRL agent is responsible to improve sequencing decisions by dynamically sorting wait queues in front of each station according to an appropriate dispatching rule. Second, assignment decisions are improved by assigning tasks to another station or another worker. A dual-discrete action space was implemented for this purpose. It consists of two subsets $AS_1, AS_2$ from which one action is selected. $AS_1$ provides a set of selectable dispatching rules such as Slack Time Rule (STR). $AS_2$ offers actions to change task assignments after the dispatching rule has been applied. For example, the agent then recognizes, that two jobs with an early due date are competing for processing, the task with the second highest priority is shifted to another station to resolve the bottleneck (station flip $SF$). With a worker flip ($WF$), the task with the highest priority is shifted to another worker who has the capability to process the task. As a third option, the assignment is not changed (no flip $NF$). The state space was specified to be as generic as possible by applying wait queue metrics such as work in progress (WIP) and throughput rates from the law of [44] (see Appendix A).

$$AS_1 = \{SPT, LPT, MTWR, STR\} \quad (24)$$

$$AS_2 = \{SF, WF, NF\} \quad (25)$$

The reward is divided into an intermediate reward $IR$ and a final reward $FR$ at the episode end. The final reward is calculated according to the overall multicriteria objective value achieved by the agent $fit_{achieved}$, which is subtracted from the sample label $fit_{label}$ and multiplied by a large number considering the number of episode steps $s$. This ensures that the final reward is given a high priority. The $IR$ is calculated after each step and is used for short-term conditioning by improving important key figures such as mean throughput rate or mean slack time remaining (see Appendix A).

$$FR = (fit_{label} - fit_{achieved}) * 20s^2 \quad (26)$$

The DRL model was implemented in the Stable-Baselines3 framework (see [45]), which provides a standardized programming interface to implement DRL algorithms with PyTorch. Here, we implemented a proximal policy optimization (PPO) algorithm, which is a well-known and performant policy gradient algorithm with actor-critic DNN [46], that has often been used for various cases and benchmarks in recent literature. With some exceptions such as the network architecture, the learning rate or the discount factor, we used the standard hyperparameters of the Stable-Baselines3 PPO implementation (see Appendix B). The training process and DRL injection to the memetic algorithm are described in Section V.

## V. COMPUTATIONAL STUDY

In this section, the proposed scheduling framework is evaluated and discussed within numerical experiments. Since it is an applied research project and our industrial partner required equally weighted objectives that do not prioritize one metric, MS and TT have equal weights in our investigation. Thus, all experiments were performed for objective weights $w_1 = 0.5$ (MS) and $w_2 = 0.5$ (TT). Self-generated benchmark data and real-world data from our partner have been used in the experiments. The following questions were answered:

1) How runtime intensive is an exact method? (see Section V-A)
2) How can suitable test data be generated and what is the structure of the real-world data? (see Section V-A)
3) How successful is a DRL agent in learning a policy? (see Section V-B)
4) How good are the results of the framework compared to traditional methods? (see Section V-C)
5) Which runtime effects can arise with parallelization and what are the conclusions for a reactive use of the framework? (see Section V-D)

### A. DATA GENERATION AND COMPLEXITY ESTIMATION

This section relates to the first two questions. First, the computational intensity of the exact method has been tested. We used an IBM CPLEX V12.8 solver in a high-performance environment (see Appendix B). It could be observed that the problem instance structure had a large impact on the computing time. Even with small inputs (e.g. 5 jobs), small variations could lead to an exorbitant increase in computing time. The complexity has increased especially when the job structure or the resources have been varied. Consequently, the calculation process was aborted after several hours and it can be concluded that an exact method is not suitable to be utilized in practice. Even in a high-performance environment, the NP complexity is too high for solving large inputs with many jobs or resources. Since the JSSP is already NP-hard [2], an even greater complexity can be conjectured for the proposed problem.

Due to the novelty of the developed MILP and the need to assess different heuristic methods, it was necessary to generate dedicated test data with a suitable level of complexity. The test data was then validated using the CPLEX solver: datasets that are too easy (too fast) to solve, as well as datasets that are too difficult to solve, were excluded. A calculation time of approximately 10 minutes to find the global optimum ensured that very efficient heuristics are challenged, while maintaining the feasibility of obtaining global optima or good local optima in a reasonable amount of time. Thus, this time bound represents a suitable balance between sufficient difficulty and solvability for the heuristics.

Two synthetic datasets with different focuses were generated: The first dataset *GBRT01* has a greater complexity in the job paths, but fewer assignable resources. On the other hand, the second dataset *GBRT02* has a low job structure but more operations and more allocable resources (see Figure 4). A multicriterial global optimum was found for both datasets (see Table 3). Furthermore, we used feasibly solvable real-world datasets *GBRT03-08* based on typical planning situations from our industrial partner. Nevertheless, due to the job and resource structure, less complexity is given. For example, the worker skills are clearly defined by a capability matrix and not every worker can work at every station. In addition, there are not that many flexible stations. With the real-world data, one time unit in the scheduling method corresponds to

TABLE 3: Benchmark datasets used (GBRT01-02: Synthetic data, GBRT03-08: Real-world data from our industry partner). GBRT03-08 global optimum could not be computed in a reasonable time. The global optimum is represented as MS/TT (both objectives considered synchronously with equal weights).

| GBRT | Jobs/tasks | Stations/workers | Global optimum |
| --- | --- | --- | --- |
| 01 | 6/14 | 2/2 | 728/83 |
| 02 | 3/10 | 6/3 | 325/75 |
| 03 | 112/251 | 20/24 | ? |
| 04 | 61/89 | 13/23 | ? |
| 05 | 61/144 | 12/23 | ? |
| 06 | 94/188 | 20/24 | ? |
| 07 | 73/143 | 15/24 | ? |
| 08 | 84/168 | 15/23 | ? |

30 minutes in reality. However, a global optimum could not be computed. The exact computing time is unclear.

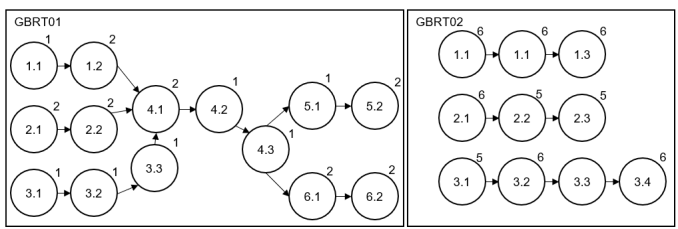

FIGURE 4: GBRT01-02 job paths visualized as DAGs. GBRT01 consists of 6 highly structured jobs with 2-3 tasks each. Job 1 task 1 (1.1) can be assigned to one machine, while the successor task 1.2 can be flexibly assigned to two machines. GBRT02 consists of 3 jobs with 3-4 tasks each but with a lower DAG structure. However, there are more flexible stations to choose from. GBRT03-05 cannot be displayed due to the large number of planning objects. More details about the data structures can be seen in the public code.

### B. DRL TRAINING

This section relates to question 3: It was examined how the DRL agent behaves during training and how it learns a policy in order to make situationally appropriate sequencing or assignment decisions. The training process had the following steps: First, a starting solution for an input problem instance was provided as a sample. It can be generated, for example, using a shortened form of the memetic algorithm (e.g. 10 generations). Due to the specialization in one problem instance, the training could be carried out in seconds to a few minutes. Since we do not claim to develop a general agent for all kinds of inputs, temporary overfitting to a specific scenario was accepted as an inherent aspect of this online learning method. In the subsequent training, the DES was used as the environment with which the agent interacts. The agent starts the simulation and waits until a decision has to be made. Looking at Figure 3, a decision situation arises when a station's queue is to be sorted. In this case, the simulation pauses and the current state is given to the agent.

Based on the state, the agent now determines a dispatching rule and if a resource flip has to be carried out, which activates the simulation again. The DRL hyperparameters used can be found in Appendix B. Figure 5 shows the results obtained from a PPO training processes using the example of GBRT03. During training, a clear trend of decreasing loss and increasing reward emerged. In the end, both metrics eventually converged, which is an indicator of successful training and reaching a stable policy. The overall negative end reward is a consequence of exploration that also took place late in the training. Due to the quadratic function for calculating the final reward, the agent receives high penalties in a few cases. Nevertheless, the figures indicate that the DRL agent has learned an appropriate policy to apply sequencing and assignment decisions for the DRC-FJSSP. The next section analyzes how the agent can be injected into the higher-level metaheuristic to improve its results. This also includes conclusions about how the learned policy is applied.

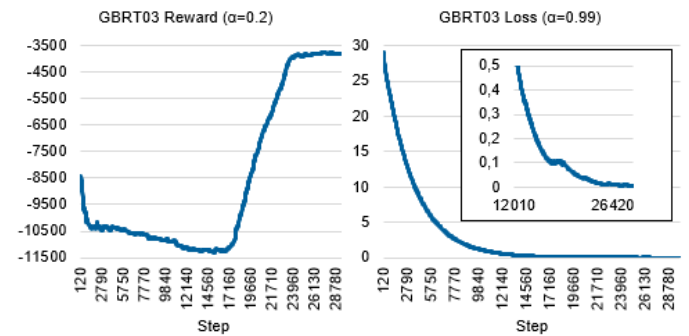

FIGURE 5: PPO agent loss and reward exemplary for GBRT03 (30,000 training steps). Similar patterns could be observed for GBRT01 and GBRT02.

### C. ALGORITHM BENCHMARK

In this experiment, the scheduling framework was analyzed in terms of scheduling performance (see question 4). The schedules generated with the conventional approaches were compared with those of the GASA+RL approach, where the trained actor critic DNN is injected to the memetic algorithm.

#### 1) Experiment setup

For representative benchmarks, we used a modern GA with adaptive hyperparameters as a basis [47]. The hyperparameters can be found in Appendix B. For GASA, we utilized an implementation of SA with restart strategy (SARS) based on Yu *et al.* [48]. Moreover, standalone SARS, the well known Tabu Search (TS) method and the dispatching rules *Most Total Work Remaining (MTWR)* and *Slack Time Rule (STR)* were tested. When using the dispatching rules for sequence decisions, the tasks were distributed evenly over the resources. Due to various hardware-related runtime effects such as efficiency advantages through parallelization, the metaheuristics were compared based on their actual complexity. For better comparability, the number of operations for creating and evaluating neighborhood solutions was used instead of the physical runtime. All metaheuristics have a runtime complexity of $O(n)$ where $n$ is the number of operations and evaluations until the stopping criterion is reached. $n = 500$ was set for all algorithms so that their logical performance could be compared. However, Sect. V-D deals with further runtime optimization that is possible through parallelization. An agent was trained for each of GBRT01, GBRT02, and GBRT03. No separate agent was trained for the additional real-world datasets. Instead, the agent from GBRT03 was used, as only the input changed and not the fundamental scenario. Consequently, the experiments for GBRT04-08 also examine the generalizability of the agent.

TABLE 4: Achieved MS and TT mean values for datasets GBRT01-08 by the different heuristics STR, MTWR, TS, SARS, GA, GASA and GASA+RL. The proposed approach GASA+RL outperforms traditional methods in most metrics. All metaheuristics have the same logical runtime (500 operations and evaluations to create neighbor solutions). The physical runtime may vary depending on the dataset and due to various hardware effects (approx. 30-150 seconds).

| Dataset (GBRT) | STR | MTWR | TS | MS / TT SARS | GA | GASA | GASA+RL |
|---|---|---|---|---|---|---|---|
| 01 | 1076.3 / 436.2 | 1091.8 / 519.9 | 774.9 / 191.7 | 777.5 / 179.9 | 748.7 / 190.8 | 735.1 / 180.8 | **730.0 / 172.2** |
| 02 | 438.6 / 346.6 | 434.0 / 333.3 | 340.1 / 113.8 | 336.2 / 112.2 | 336.8 / 92.3 | 335.0 / 89.1 | **333.0 / 83.7** |
| 03 | 66.9 / 15017.7 | 231.1 / 16449.4 | 203.6 / 11299.9 | 196.2 / 10723.0 | 188.1 / 10753.1 | **186.3** / 10538.7 | 189.8 / **10199.9** |
| 04 | 53.6 / 3209.6 | 52.95 / 3336.9 | 52.25 / 2727.5 | 52.35 / 2761.6 | **52.0** / 2742.8 | **52.0** / 2740.9 | **52.0 / 2715.7** |
| 05 | 107.6 / 4852.8 | 112.2 / 5378.4 | 93.6 / 4000.5 | **93.5** / 4014.9 | 94.3 / 4030.5 | 93.9 / 4002.8 | 93.6 / **3947.35** |
| 06 | 241.2 / 9612.8 | 212.9 / 10704.7 | 179.5 / 7268.9 | 175.7 / 7380.3 | 175.2 / 7212.8 | 175.1 / 7282.7 | **174.6 / 7020.1** |
| 07 | 145.9 / 6358.2 | 139.7 / 6894.8 | 122.9 / 4678.2 | 117.3 / 4769.1 | 115.4 / 4746.3 | 114.7 / 4695.9 | **114.2 / 4668.1** |
| 08 | 141.8 / 8040.4 | 135.4 / 8716.6 | 115.8 / 6303.7 | 115.7 / 6288.5 | 115.0 / 6244.0 | 115.5 / 6169.9 | **114.7 / 6162.8** |

TABLE 5: Achieved MS and TT standard deviations (addition to Table 4)

| Dataset (GBRT) | STR | MTWR | TS | MS / TT SARS | GA | GASA | GASA+RL |
|---|---|---|---|---|---|---|---|
| 01 | 95.9 / 89.5 | 91.1 / 95.1 | 44.9 / 43.6 | 45.8 / 24.6 | 19.2 / 19.3 | 19.6 / 23.6 | **17.2 / 17.8** |
| 02 | 57.7 / 120.2 | 56.6 / 125.8 | 8.4 / 31.7 | 9.2 / 40.5 | 8.8 / 11.7 | 6.9 / 12.7 | **5.0 / 5.3** |
| 03 | 41.5 / 891.2 | 33.4 / 1142.5 | 19.3 / 902.1 | 10.3 / 814.2 | **2.3 / 288.7** | 2.7 / 320.6 | 2.5 / 332.5 |
| 04 | 3.5 / 55.3 | 1.9 / 54.9 | 0.9 / 47.5 | 0.7 / 51.2 | **0.0** / 32.6 | **0.0** / 28.5 | **0.0 / 12.3** |
| 05 | 6.6 / 152.1 | 11.0 / 222.9 | 1.7 / 166.0 | **1.6** / 170.3 | 2.4 / 127.1 | 1.8 / 126.8 | **1.6** / **69.9** |
| 06 | 40.1 / 643.2 | 37.4 / 805.8 | 16.6 / 426.7 | 4.8 / 299.3 | 2.7 / 336.7 | 2.5 / 401.4 | **1.9 / 290.7** |
| 07 | 15.7 / 450.9 | 19.5 / 528.6 | 15.9 / 271.5 | 7.5 / 146.8 | **4.5** / 147.3 | 5.5 / **113.9** | **4.5** / 191.1 |
| 08 | 14.3 / 364.9 | 14.4 / 406.5 | 2.5 / 427.7 | 3.4 / 262.5 | 2.2 / 270.5 | 2.7 / 293.6 | **1.3 / 227.0** |

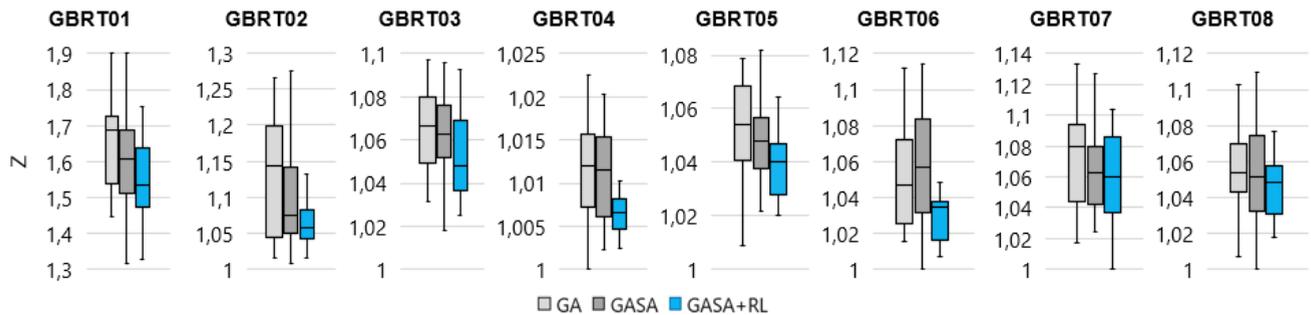

FIGURE 6: Result comparison between GA, GASA and GASA+RL for GBRT01-08 considering the achieved multicriterial objective value $Z$ (see Eq. 22). On average, GASA+RL generates the best schedules and clearly outperforms GA and GASA.

2) Results and discussion

Table 4 and Table 5 show the summary statistics for multicriteria objectives achieved considering mean values and variability. The results indicate that dispatching rules have a poor performance but are easy and quick to compute. Nevertheless, the rules are suitable for generating an initial feasible solution for subsequent search methods. Utilizing the trajectory methods TS and SARS resulted in significantly better outcomes, with SA marginally outperforming TS. Moreover, even better local optima were discovered using population-based approaches (GA, GASA, GASA+RL). This can be explained by the fact that population-based methods have a better ability to explore the breadth of the solution space. Consequently, due to the many possible combinations in the optimization model and input data as well as by the high standard deviations, the solution space appears to consist of many local optima.

For a better overview of the population-based methods, Figure 6 displays related statistical metrics in form of a box chart considering the scalarized objective value $Z$ (see Eq.

22). The first remarkable aspect of the results is the different performances on GBRT01 and GBRT02. While for GBRT02 consistently good fitness values near to the global optimum were achieved, the pattern did not appear for GBRT01. Even the best results still had a distance of more than 30% to the global optimum. Although relatively good MS values were achieved, this resulted in a deterioration in TT. Overall, these results indicate that the proposed memetic algorithm is well suited to schedule production processes with many flexible resources. With complex job paths, it is difficult to find good results quickly. This applies in particular to the TT minimization. A possible explanation for this issue might be that setup operations can be left shifted and start before the processing operation or even before the job is ready. In the case of non-sequential job paths with many predecessor/successor relationships, few resources and sequence-dependent setup times, it can thus be suggested that it is a difficult decision, when to set up a machine in advance.

What is also striking about the table and figures is the dominance of GASA over GA and GASA+RL over GASA in most cases. The average superiority of GASA over GA was to be expected since similar outcomes had already been shown in many other studies. However, this confirms the validity of the memetic algorithm implemented and the hyperparameters chosen. Interestingly, the GASA results had a comparable high standard deviation, which is especially observable in the TT case. A possible explanation for this might be that GASA sometimes finds better schedules due to greater exploitation through SARS, which is also reflected in the minimum values (see outliers in the box chart).

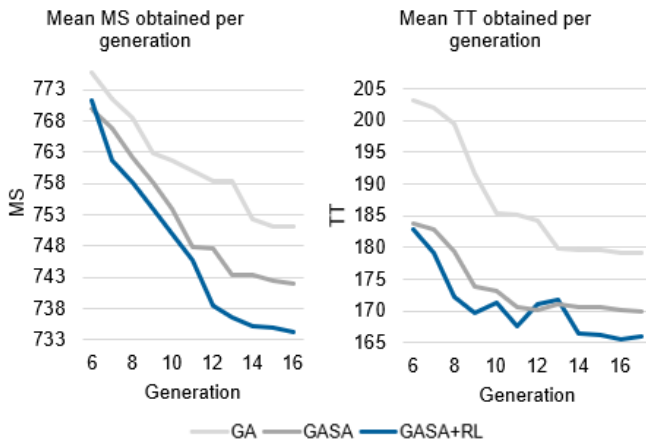

FIGURE 7: Comparison between GA, GASA and GASA+RL in early iterations considering MS and TT. GASA+RL outperforms conventional approaches and identifies better local optima early in the process (exemplary for GBRT01).

The dominance of GASA+RL is the most interesting result. Regarding the $Z$ score visualized in the box chart, better average, maximum, Q1 and Q3 values could be achieved in the most cases. Especially, the model's predictions may prevent random upward outliers and result in comparatively low Q3 values. This is an important finding and an indicator of a good reliability with relatively short runtimes. Looking at Table 4, it is noteworthy that GASA+RL achieves significantly better TT values for the real-world data (GBRT03-08), but not so in the MS case. This result may be explained by the fact that the agent prioritized the TT optimization due to possible better robustness and predictability while learning the policy. This effect could not be observed in the synthetic datasets (GBRT01-02) and is probably due to the structure of the real-world data. For a further insight, Figure 7 shows how the different approaches already differ in early iterations. Across all approaches, the greatest advances are made in about the first 10 generations. Here, GASA+RL already achieves significantly better fitness values than GA and GASA. This indicates that DRL improves the discovery of local optima early in the process.

The experiment confirms that a suitably trained DNN fed into the metaheuristic leads to better decisions than a randomized creation of neighborhood solutions as it is common for metaheuristics. The prerequisite is that the agent has learned a generally applicable policy that is suitable for the dataset. Based on the hypothesis that conventional metaheuristics can be improved in this way, further research should be undertaken to investigate DRL-based hybrid methods and their applicability to other optimization problems. Related to the production process optimization, further work is required to analyze the suitability of multi agent concepts for DRC-FJS. For example, separate agents for sequencing, machine and worker assignment decisions could be examined in terms of their performance and generalizability. In this experiment, it could be shown that the neural network trained for GBRT03 also performs well for the other real-world datasets GBRT04-08. In completely different scenarios, however, the agent's performance is poor. Thus, the perhaps most important question is, how DRL agents can internalize a policy as generally as possible for different manufacturing scenarios.

### D. PARALLELIZATION EFFECTS

The purpose of this experiment was to investigate the effect of parallelizing the memetic algorithm on multiple computing processes (question 5). With a view to operational practice, a short algorithm runtime is an important requirement. On the one hand, this refers to the reactive use of a planning method where a schedule must be quickly regenerated when unplanned events occur. On the other hand, the method must be efficient so that extensive realistic scenario analyses can be performed dynamically. By choosing a population-based method, we were able to achieve this in a targeted manner. As described in Sect. IV-A, the population of each generation is splitted into equally sized batches, where each batch is processed in a parallel computing process. In this way, it is possible to distribute the operations of the memetic algorithm (crossover, mutation, SA exploitation) as well as the computationally intensive simulation-based evaluation over the computing resources. Figure 8 provides mean values for the runtime per algorithm iteration. There was a significant correlation between mean iteration time and the number of parallel processes: The speed of the procedure can be optimized by parallelization without affecting the resulting quality. In this way, a schedule can be generated in seconds to a few minutes, depending on the input size and hardware. This outcome further supports the idea of providing the proposed algorithm as a scalable cloud-based smart service. In modern managed cloud environments, an abundant number of processes can be dynamically spawned based on demand, allowing a free allocation of computing resources.

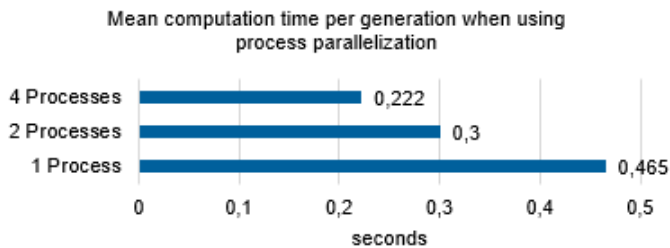

FIGURE 8: Computation time per algorithm iteration without and with parallelization (2 and 4 processes). Parallelization with 4 processes improves the runtime by nearly half the runtime taken with no parallelization for GA, GASA and GASA+RL.

### E. LIMITATIONS AND IMPLICATIONS

Based on the previous experiments, the essential shortcomings and needs for future works are summarized. The proposed method enables deterministic scheduling without uncertain parameters or stochastic events. The objectives MS and TT are balanced with equal weights and there is no analysis on how to set the weights or how the algorithm performs with different weights. Through the conception as a simheuristic, the use case for dynamic scheduling or re-scheduling policies can be extended. Worker skills and skill-dependent operation times can be used for workforce planning. However, no further effects of human work are currently taken into account (e.g. preferences for operation types, motivation, mental stress, exhaustion or physical limitations). Short-term lot size planning to minimize setup times is possible. Long-term dynamic lot size planning taking into account primary requirements in subsequent periods is not possible. This study considers an application and not an in-depth investigation of DRL. Due to very different possible inputs, we were not able to find a general DRL hyperparameter setting for all scenarios. This applies in particular to the number of training steps and DNN update intervals. This is not a disadvantage of this study, as we were still able to confirm that DRL can improve the results of a memetic algorithm. However, further research should be undertaken to examine the predictability of appropriate hyperparameters for different production environments considering varying numbers of stations, workers or tasks.

Thus, a central scientific question raised by this study is how DRL can be better generalized to solve varying and complex planning scenarios. In terms of meta-reinforcement learning [49], a highly generic agent may be able to replace meta-heuristics, as a pre-trained neural network would have access to existing knowledge and would not need to completely recalculate every instance. This is an crucial potential to save computing time. There is also the question of how proactive planning approaches can be integrated into the scheduling framework, taking uncertainties such as dynamic events into account. In the future, field studies should be carried out to examine the effectiveness of holistic, predictive and reactive scheduling algorithms and how these can be successfully integrated into business processes. Moreover, it would be interesting to assess the effects and feasibility of adding more resource types and real-world processes. This refers to continuous production, inventory levels, special tools, machine buffers, tasks involving multiple workers, dynamic events and customer demands as well as transportation logistics such as the consideration of conveyor belts.

From a managerial perspective, the results of this research support the idea that human-centered scheduling in complex and flexible production processes can be optimized using modern algorithms and technologies such as generic smart services or digital twins. The findings will be of interest to several companies in that area to reduce intransparency, stress and manual effort in terms of reactive scheduling. The deployment of the framework can play a crucial role for a realistic production planning in the domain. An important prerequisite for this is a consistent data basis from operational systems containing all parameters required by the proposed optimization model.

### VI. CONCLUSION

This study set out to develop a novel memetic algorithm enriched with DRL to solve an extensive DRC-FJSSP. The paper's motivation was to address a shortage of prior extensive scheduling models as well as efficient, generic heuristics. The contribution was an algorithm framework capable to map these complex requirements and reliably generating schedules with minimal computing time.

First, a MILP was induced based on practical requirements of human-centered manufacturing processes. Then, a scheduling framework was proposed and tested with real-world data for simultaneous MS and TT minimization. The framework integrates a freely expandable DES and a novel value encoding, which enables both job shop and flow shop organizations. Within a computational study we were able to confirm that a memetic algorithm with DRL outperforms random-based metaheuristics. DRL enables the substitution of a randomized neighborhood creation through targeted sequencing and assignment decisions. As a result, better and more reliable schedules could be generated in fewer iterations. With parallel computing, the algorithm efficiency could be further increased so that the method can be used reactively and in a scalable manner. Under the hardware used and depending of the input dataset, the schedules could be generated in a few seconds to a few minutes.

The developed framework enables our industry partners to plan more holistically compared to conventional methods such as job shop scheduling or critical path method. By considering human workers and their capabilities as well as sequence-dependent setup times, a short-term workforce and lot size planning is synchronized with job scheduling.

.

## APPENDIX A  DRL STATE SPACE AND REWARD FUNCTION

**Algorithm 1** Intermediate reward calculation (pseudocode snippet)

1: $Input: C, L, a_1, a_2$  ▷ current and last state features, actions
2: $r \leftarrow 0$
3: **if** $a_2 = SF \wedge L_8 = 0$ **then**
4:     $r \leftarrow -3$
5: **else if** $a_2 = SF \wedge L_3 > L_5$ **then**
6:     $r \leftarrow 2$
7: **else if** $a_2 = WF \wedge \frac{L_6}{L_9} > 1$ **then**
8:     $r \leftarrow 1$
9: **end if**
10: **if** $a_1 = STR \wedge L_{15} < L_{16}$ **then**
11:     $r \leftarrow r + 1$
12: **end if**
13: **if** $C_{12} > L_{12} = 0$ **then**
14:     $r \leftarrow r + 3$
15: **end if**
16: **if** $C_{14} > 0$ **then**
17:     $r \leftarrow r + 3$
18: **end if**
19: **return** $r$

TABLE 6: DRL state space features (i: Feature index)

| i | Description |
|---|---|
| 1 | Relative environment time: Current simulation time in relation to the average WIP per station. A relative time greater than 1 is an indicator of wait times and that work is not evenly distributed across resources. |
| 2 | Production stage: Average topology group $G$ of available tasks at the station. This feature is intended to describe the position at which the station is located in the production line. |
| 3 | Relative station WIP: Duration of all tasks that have not yet been completed (in relation to the overall WIP). This feature provides information on how much work still needs to be done by the station. It is intended to be an indicator of potential congestions. |
| 4 | Mean successor station WIP: The average duration of the tasks that have not been completed across all stations that follow directly in the production line. The intention here is to make accumulations and bottlenecks between related stations recognizable. |
| 5 | Mean station WIP: Average duration of non-completed tasks over all stations. |
| 6 | Relative worker WIP: Duration of all tasks that have not yet been completed for the workers at the station. This is intended to give an indication of whether comparatively much or little human labor is required at the station. Thus, it describes the automation degree. |
| 7 | Number of available tasks at the station, that are directly processable (relative to all processable tasks over all stations) |
| 8 | Competing tasks (binary encoded): 1, if more than one task is currently processable at the station. |
| 9 | Number of processing slots at the station. This feature describes the situational capacity of the station, how many tasks can be processed in parallel and how many workers can work here at the same time. |
| 10 | Current station throughput rate. According to Little's Law, this feature should be used in combination with the WIP information to infer waiting times, which are an indicator for the MS [50]. |
| 11 | Current mean throughput rate of all successor stations |
| 12 | Current mean throughput rate over all stations |
| 13 | Current throughput rate standard deviation over all stations |
| 14 | Minimum job slack time remaining at the station |
| 15 | Mean job slack time remaining at the station |
| 16 | Mean job slack time remaining over all stations |
| 17 | Job slack time remaining standard deviation over all stations |

## APPENDIX B  SETTINGS AND HYPERPARAMETERS FOR THE NUMERICAL EXPERIMENTS

TABLE 7: GA hyperparameters

| Hyperparameter | Value |
|---|---|
| Initialization method | Equally distributed resources |
|  | Random dispatching rules |
| First population size | 50 genomes |
| Selection method | Best |
| Number of survivors | 8 |
| Offspring size | 20 |
| Mutation probability | Adaptive (linearly decreasing) |
| Combination probability | Adaptive (linearly increasing) |
| Stopping criterion | 500 operations and evaluations |

TABLE 8: MILP Solver environment

| | |
|---|---|
| Solver: | IBM CPLEX V12 |
| OS: | CentOS 7.5.1804 |
| vCPU: | Intel Xeon E5-2630 V4 2.2 GHz with 28 cores assigned |
| RAM: | 64 GB DDR4 2133 MHz |

TABLE 9: DRL hyperparameters for the training process and for the actor-crititic PPO network

| Hyperparameter | Value |
|---|---|
| Overall training steps | Dynamic (30,000 on GBRT03) |
| Network update interval (steps) | Dynamic (10 on GBRT03) |
| Learning rate | 0.0001 (linearly decreasing) |
| Discount factor | 0.999 |
| Activation function | Rectified Linear Unit (leaky) |
| Network architecture (hidden layers) | $2^9$ neurons (shared) |
|  | $2^7, 2^6$ neurons (value net) |
|  | $2^6$ neurons (policy net) |

## NOTES ON CONTRIBUTORS

**Felix Grumbach**: Conceptualization, Methodology, Software, Validation, Formal analysis, Investigation, Data Curation, Writing - Original Draft, Writing - Review & Editing, Visualization, Project administration. **Nour Eldin Alaa Badr**: Investigation (related work analysis), Data Curation, Writing - Original Draft (literature review), Writing - Review & Editing. **Pascal Reusch**: Resources, Funding acquisition, Project administration, Supervision. **Sebastian Trojahn**: Supervision.

## CODE AVAILABILITY

Open source code is publicly accessible at https://doi.org/10.17605/OSF.IO/JRVFC.

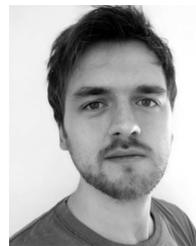

FELIX GRUMBACH recieved the B.Sc. degree in information systems from Bielefeld University of Applied Sciences (Germany) in 2015 and the consecutive M.Sc. degree from University of Hagen (Germany) in 2020. He is currently a research associate at Center For Applied Data Science (Bielefeld University of Applied Sciences) and a Ph.D. student at the Doctoral Center for Social, Health and Economic Sciences (Saxony-Anhalt, Germany). Mr. Felix Grumbach's research focuses on the robust and holistic optimization of complex production processes with the help of Machine Learning techniques.

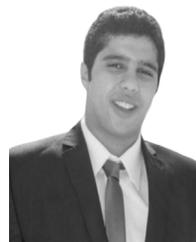

NOUR ELDIN ALAA BADR recieved the B.Sc. degree in electrical engineering from Higher Technological Institute (Cairo, Egypt) in 2016 and the M.Sc. degree in intelligent systems from Bielefeld University (Germany) in 2022. He is currently a research associate at Center For Applied Data Science (Bielefeld University of Applied Sciences). Mr. Badr focuses on researching and developing intelligent production scheduling systems.

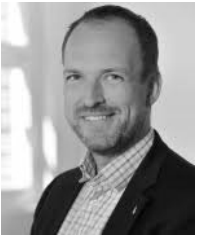

PASCAL REUSCH has a Ph.D. in economics and is a professor of production and industrial management at Bielefeld University of Applied Sciences. He is also a founding member of the Center for Applied Data Science (CfADS). Dr. Reusch has several years of practical professional experience in the industry.

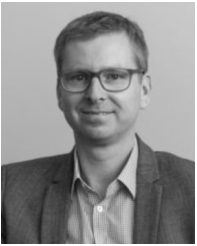

SEBASTIAN TROJAHN has a Ph.D. in engineering and is a professor of supply chain management, operations management and digital logistics at Anhalt University of Applied Sciences (Germany). He has several years of practical professional experience in the industry.

∙ ∙ ∙